\documentclass[letterpaper, 10 pt, conference]{ieeeconf} 
\usepackage{amsmath}
\usepackage{mathtools}
\usepackage{makecell, multirow}
\usepackage{graphicx}
\usepackage{caption}
\usepackage{subcaption}
\usepackage{hyperref}
\usepackage{amsfonts} 
\usepackage{cleveref}
\usepackage{bm}
\usepackage{multirow}
\usepackage{booktabs}
\usepackage{verbatim}
\usepackage{listings}
\usepackage{kantlipsum}

\let\labelindent\relax			
\usepackage{enumitem}

\begin{document}
\title{Safe and Sample-efficient Reinforcement Learning for Clustered Dynamic Environments}
\author{Hongyi Chen and Changliu Liu, \IEEEmembership{Member, IEEE}
\thanks{Manuscript received September 10, 2021; revised November 20, 2021;
accepted December 12, 2021.}
\thanks{Hongyi Chen is with the Institute for Robotics and Intelligent Machines, Georgia Institute of Technology, North Ave NW, Atlanta, GA 30332, USA (e-mail: hchen657@gatech.edu).}
\thanks{Changliu Liu is with the Robotics Institute, Carnegie Mellon University, 5000 Forbes Ave, Pittsburgh, PA 15213, USA (e-mail: cliu6@andrew.cmu.edu).}}

\maketitle
\thispagestyle{empty}

\begin{abstract}

This study proposes a safe and sample-efficient reinforcement learning (RL) framework to address two major challenges in developing applicable RL algorithms: satisfying safety constraints and efficiently learning with limited samples. To guarantee safety in real-world complex environments, we use the safe set algorithm (SSA) to monitor and modify the nominal controls, and evaluate SSA+RL in a clustered dynamic environment which is challenging to be solved by existing RL algorithms. However, the SSA+RL framework is usually not sample-efficient especially in reward-sparse environments, which has not been addressed in previous safe RL works. To improve the learning efficiency, we propose three techniques: (1) avoiding behaving overly conservative by adapting the SSA; (2) encouraging safe exploration using random network distillation with safety constraints; (3) improving policy convergence by treating SSA as expert demonstrations and directly learn from that. The experimental results show that our framework can achieve better safety performance compare to other safe RL methods during training and solve the task with substantially fewer episodes. Project website: \href{https://hychen-naza.github.io/projects/Safe\_RL/}{https://hychen-naza.github.io/projects/Safe\_RL/}.
\end{abstract}

\section{INTRODUCTION}
Recently, reinforcement learning (RL) shows promising results in a series of artificial domains; but it's still challenging to develop applicable RL algorithms for a real system due to nine challenges discussed in \cite{49999}. Our paper focuses on the two challenges: satisfying safety constraints and learning from limited samples. 

Ideally, 0-safety violation should be guaranteed during both training and execution as failures are expensive and dangerous. A comprehensive survey on optimization-based and exploration-based safe RL methods can be found in \cite{JMLR:v16:garcia15a}. Within the optimization-based methods, they redesign the reward function, change the learning objective and add soft constraints to balance the reward and the risk \cite{10.5555/1622519.1622522, ray2019benchmarking, HEGER1994105,DBLP:journals/corr/AchiamHTA17, 10.5555/3042573.3042784}; however, no safety guarantee can be derived from these methods. Within the exploration-based methods, they modify the exploration process to avoid risky situations by incorporating external
knowledge \cite{ApprenticeshipLearning, lfh} and using a risk-directed exploration \cite{smartexploration}. The probability-based methods construct shields to avoid violating the safety constraints, but they only meets the safety requirement with some probability and are hard to be generalized to continuous systems \cite{hasanbeig2020cautious,Jansen2018ShieldedDI}. Control barrier function (CBF) method is used to provide hard constraints for RL and achieve 0-safety violation in simple environments like car following and lane changing \cite{article}\cite{9147584}. To handle the intersection scenario in autonomous driving, generalized control barrier function (GCBF) is used to reduce the constraints violation \cite{ma2021model}.

People evaluate the sample efficiency by measuring the amount of data necessary to achieve a certain performance threshold \cite{49999}. But collecting sample data in real world is time-consuming and expensive, and RL agents may converge to local optima when the reward is sparse and never reach the performance threshold. In a word, the sample efficiency challenge makes it hard to deploy RL algorithms quickly in real world systems. Exploration methods, like adding parameter space noise (PSN) \cite{DBLP:journals/corr/PlappertHDSCCAA17} and using random network distillation (RND) \cite{DBLP:journals/corr/abs-1810-12894} can help to solve sparse-reward problem but it is risky to explore freely in clustered dynamic environment as the systems would fail or break before learning the optimal controller. Leveraging expert demonstration data can accelerate the agent's learning \cite{DBLP:journals/corr/HesterVPLSPSDOA17}, however people usually get suboptimal demonstration as the expert demonstration is hard to access. Recent model-based deep RL approaches show a lot of promise for improving sample efficiency, however, an imperfect dynamics model can degrade the performance of the learning algorithm and lead to suboptimality \cite{DBLP:journals/corr/abs-1807-01675}.

In this work, we aim to design a reinforcement learning framework that can learn safely and efficiently even in \textit{clustered dynamic environments}. We use the safe set algorithm (SSA) \cite{liu2014control} to ensure safety. SSA has similar structures as CBF, both of which belong to energy function-based safe control \cite{9029720}. These methods can safe guard any reinforcement learning algorithm, but they can't help to find optimal policies directly, and sometimes make the system behave overly conservative. Thus, we first adapt the projection direction of SSA to generate more efficient control when possible. Also by combining SSA with normal exploration strategies, we can transform these originally unsafe explorations into safe explorations. Moreover, since getting safe expert demonstration is difficult in the real world, we decide to learn from the safe control generated from SSA online to speedup training. The key contributions of this paper are summarized below:
\begin{itemize}
    \item We propose the SSA based safe RL training framework and prove this framework can guarantee safety with high probability even in clustered dynamic environments, except the cases that no safe control exists.
    \item We propose three techniques to improve the learning efficiency: adapting the SSA, exploring under safety constraints and learning from SSA demonstration. The numerical results show that we can solve the task with substantially fewer episodes and interactions.
\end{itemize}

\section{Problem Formulation}
\textbf{Environment Dynamics}\;\;\;The 2D environment contains multiple dynamic obstacles, every obstacle evolves as $\dot{x}_E = f_E(x_E, u_E)$, where the function $f_E$ represents double integrator dynamics, state $x_E$ involves the position and velocity of the obstacle and control $u_E$ represents its acceleration, which is uniformly distributed on the predefined interval.

\textbf{Robot Dynamics}\;\;\;Let $x_R \in X \subset R^{n_x}$ be the robot state containing positions and velocities in x, y axis; $u \in U \subset R^{n_u}$ be the control input, which is the accelerations in x, y axis. The robot dynamics are defined as:
\begin{equation}
    \dot{x}_R = f(x_R) + g(x_R) u =: h(x_R, u)
\end{equation}
where $f(x_R) = \left[ \begin{matrix}
	0,\mathbb{I}_2;0,0
    \end{matrix}\right]x_R$, $g(x_R) =  \left[ \begin{matrix}
    0;\mathbb{I}_2
    \end{matrix}\right]$. We assume we have the ground truth form of $f$ and $g$. Given the dynamical system above, we can formulate an MDP $(X,U,\delta,r,p)$. The transition function $p : X \times U \times X \to [0, 1] $ is defined as $p(x_{t+1}|x_t, u_t) = 1$ when $x_{t+1} = x_t+h(x_t,u_t)$ and $0$ otherwise. The reward function $r$ provides positive reward if reaching the goal state $X^*$, negative penalty if collide, and zero otherwise. The discounting factor is set to $\delta =1$.

\textbf{Safety Specification}\;\;\;The safety specification requires the robot to stay in a closed subset of state space, called the safe set $X_S$. The safe set can be presented by a zero-sublevel set of a continuous and piecewise smooth function $\phi_0: R^{n_x} \to R$, i.e. $X_S = \left\{x | \phi_0(x) \leq 0 \right\}$. In our problem, $\phi_0$ is defined as $d_{min}^2 - d^2$, where $d_{min}$ is the user defined safety distance and $d$ is the distance from robot to the closest obstacle. For safety metric, we evaluate the percentage of safety violations.

\textbf{Sample Efficiency Specification}\;\;\;To evaluate the data efficiency of a particular model, we measure the amount of data necessary to achieve a certain performance threshold:
\begin{equation}
\label{eq:ssa}
\begin{aligned}
    J^{eff} = \underset{}{\min}\,|D_i| \;s.t. R(\texttt{Train}(D_i)) \geq R_{min}. \\
\end{aligned}
\end{equation}
where $D_i$ is the data used for training the RL policy and $R_{min}$ is the desired performance threshold \cite{49999}.

\textbf{Problem}\;\;\;The core problem of this paper is to achieve safe and sample-efficient RL learning in clustered dynamic environment. The learned RL policy will map the state $(x_R, x^{c}_E)$ to control $u$, where $x^{c}_E$ means the closest obstacle to the robot. For safety, we need to monitor and modify the nominal control $u$ to keep the system inside the safe set $X_S$ and achieve least safety-violations. For sample efficiency, we need to ensure RL agent wouldn't converge prematurely to a local optimum and learn the optimal controller with fewer training data.

\section{Review of the Safe Set Algorithm}
The SSA works as a safety monitor~\cite{liu2014control}, which is suitable to safe guard the RL training. The key of SSA is to define a valid safety index $\phi$ such that 1) there always exists a feasible control input in $U$ that satisfies $\dot\phi\leq -\eta\phi$ when $\phi \geq 0$ and 2) any control sequences that satisfy $\dot\phi\leq -\eta\phi$ when $\phi \geq 0$ ensures forward invariance of the safe set $X_S$ and finite time convergence to this set. The parameter $\eta$ is a positive constant that adjusts the convergence rate. Following the safety index design rule \cite{zhao2021modelfree} for collision avoidance with single obstacle, we define the safety index $\phi$ as follows:
\begin{equation}\label{eq: safety index}
\phi = d_{min}^2 - d^2 - k\cdot \dot{d}.
\end{equation}
where $\dot{d}$ is the relative velocity of robot to obstacle and $k$ is a constant factor. We add higher order term of $\phi_0$ to the base $\phi_0$ to ensure that relative degree one from $\phi$ to the control input. As proved in \cite{liu2014control}\cite{zhao2021modelfree}, this safety index $\phi$ will ensure forward invariance of the set $\phi \leq 0 \cap \phi_0 \leq 0$ as well as global attractiveness to that set.  With the valid safety index $\phi$, we project the reference control $u^r$ to the set of safe controls that satisfy $\dot{\phi} \leq -\eta\,\phi$ when $\phi \geq 0$, and $\dot{\phi}$ is expressed as
\begin{equation}
    \dot \phi = \frac{\partial \phi}{\partial x}\,f + \frac{\partial \phi}{\partial x}\,g\; u = 
    L_{f}\phi+L_{g}\phi\; u.
\end{equation}
We compute $\phi_{i}$ for every obstacle and add safety constraint whenever $\phi_{i}$ is positive. Also we have velocity and acceleration limits. With all constraints, SSA will solve the following optimization problem through quadratic programming (QP): 
\begin{equation}
\label{eq:ssa}
\begin{aligned}
    \underset{u\in U}{\min}\,&||u-u^r||^2 
    = \underset{u\in U}{\min}\,u^\mathrm{T}\left[ \begin{matrix}
	1&0\\
	0&1
    \end{matrix}\right]u - 2u^\mathrm{T}\left[ \begin{matrix}
	1&0\\
	0&1
    \end{matrix}\right]u^r \\
    &s.t. L_{f}\phi_{i}+L_{g}\phi_{i}~u \leq -\eta\,\phi_{i}, i=1,2...m. \\
\end{aligned}
\end{equation}

However, in clustered dynamic environment, there are situations that don't even exist safe control to guarantee safety as we will discuss later (note $\phi$ in \eqref{eq: safety index} only guarantees safety with single dynamic obstacle not multiple dynamic obstacles). Besides, low sample efficiency is a problem in vanilla SSA: the agent may require long training period or even fail to learn optimal controller when the task is complex or the environment is reward-sparse. To make it work, we need to improve the sample efficiency with the following three strategies.

\section{Methodology}
\subsection{Adapting the Safe Set Algorithm}

\begin{figure}
     \centering
     \begin{subfigure}{0.4\textwidth}
         \centering
         \includegraphics[width=0.9\textwidth]{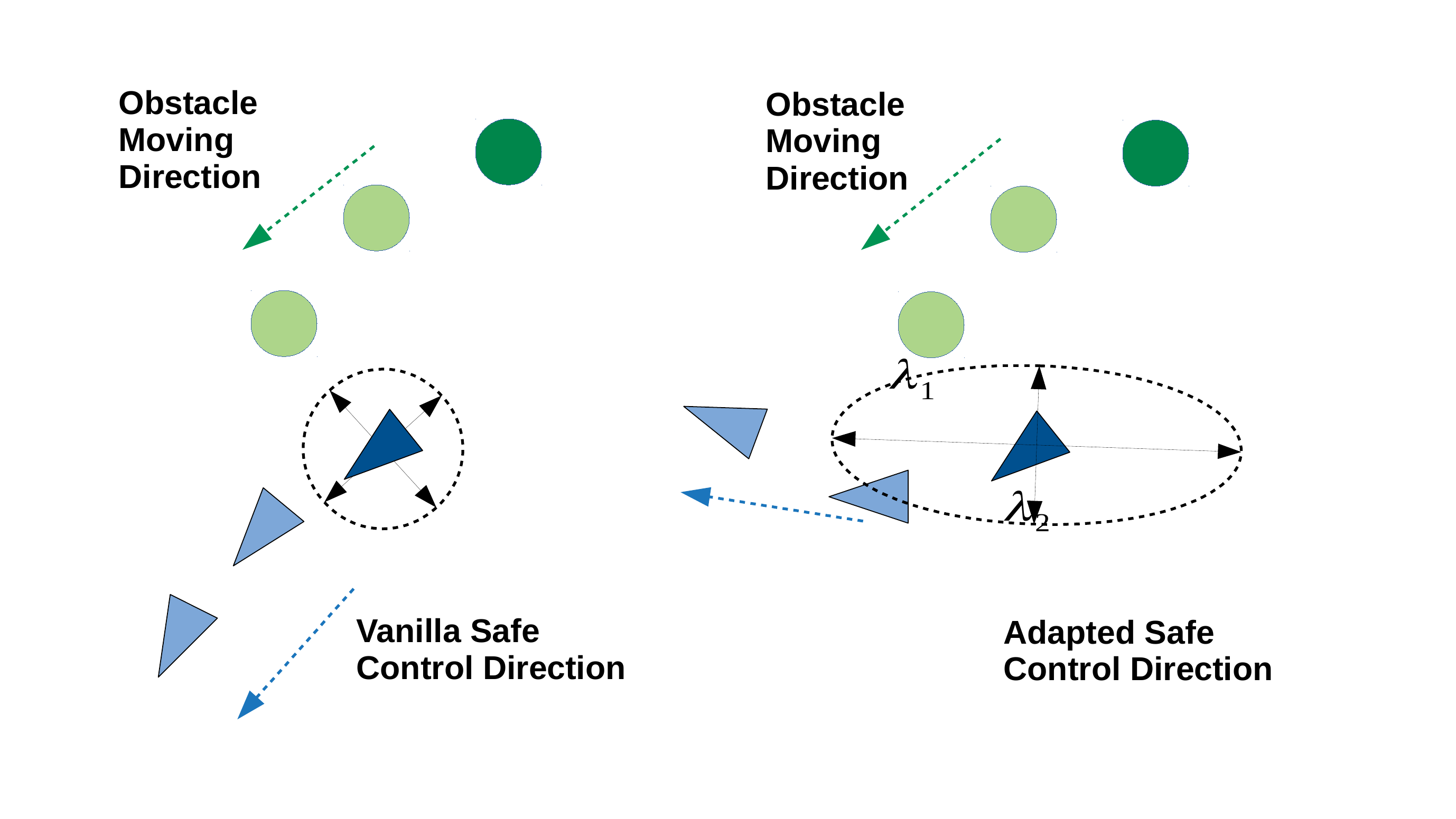}
         \caption{The robot bypasses the obstacle with adapted SSA but is pushed back by the obstacle with vanilla SSA.}
         \label{fig:adapted1}
     \end{subfigure}
     
     \begin{subfigure}{0.4\textwidth}
         \centering
         \includegraphics[width=0.9\textwidth]{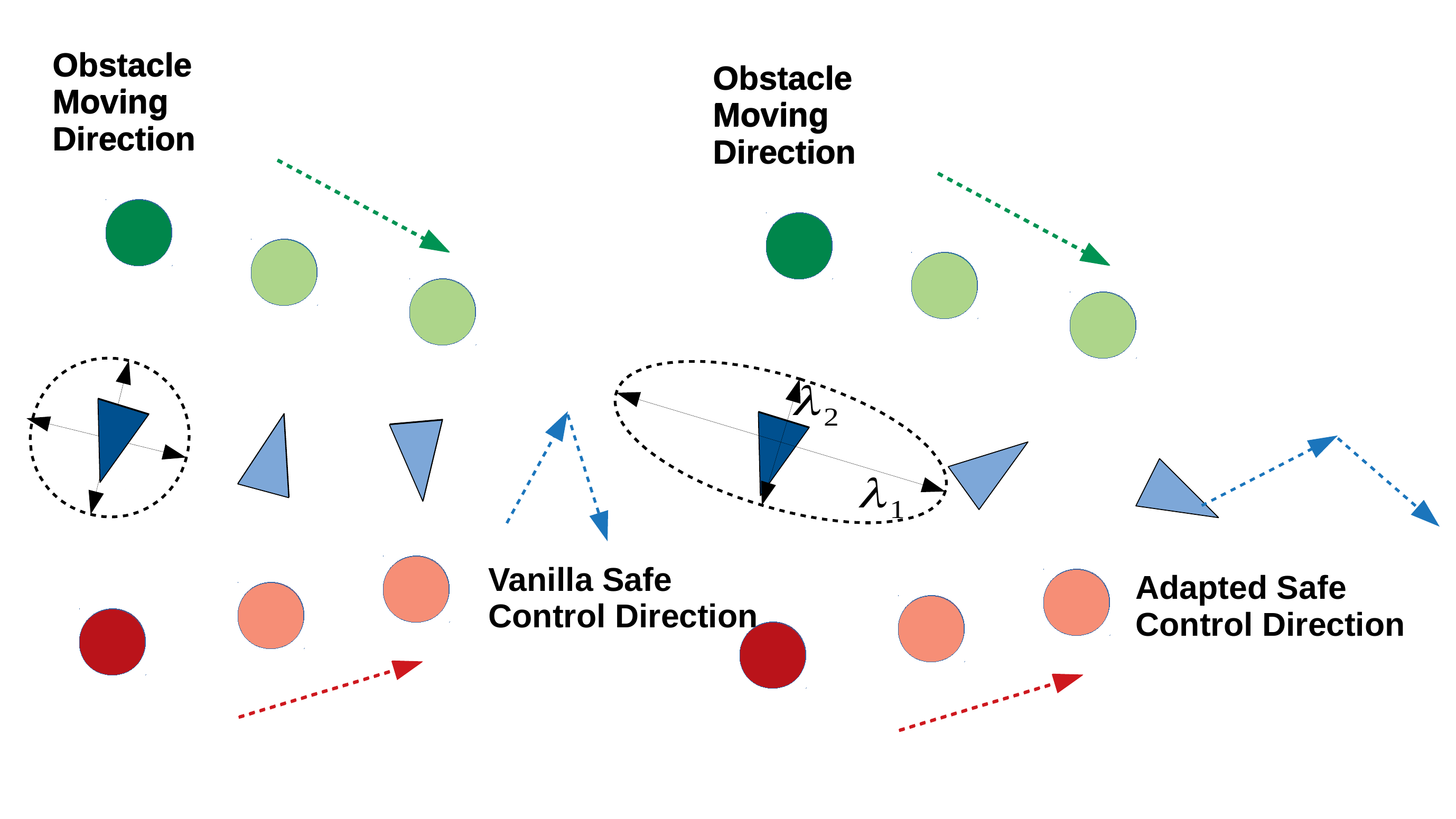}
         \caption{The robot escapes from two obstacles with adapted SSA but oscillates between them with vanilla SSA.}
         \label{fig:adapted2}
    \end{subfigure}
    \caption{Comparison between vanilla SSA and adapted SSA. The darker color presents the current positions of the robot and the obstacles. The lighter color presents the future positions of the robot and the obstacles.}
    \label{fig:adapted SSA}
    \vspace{-20pt}
\end{figure}

Vanilla SSA would output safe control that drives the system to the currently safest direction, which may not be an efficient direction in the long run, see \cref{fig:adapted1}. Besides, in multi-obstacle environment, it's not safe to directly add constraints for every dangerous obstacle whose $\phi$ is positive. In detail, since vanilla SSA only considers these dangerous obstacles, it may push the robot to the direction that is safe now but risky in the next step if there are unconsidered obstacles ($\phi$ values are negative) in that direction, see \cref{fig:adapted2}. Thus we decide to consider the current positions, estimate future positions of all approaching obstacles, and modify the direction of safe control by tuning parameters in the QP problem \eqref{eq:qp}. After adapting the projection direction, we expect to generate control signal that is safe to all approaching obstacles, even these $\phi$ values are negative, and efficient for the longer time horizon. 

\begin{equation}
    \underset{u\in U}{\min}\,||u-u^r||_{Q} 
    = \underset{u\in U}{\min}\,u^\mathrm{T}\left[ \begin{matrix}
	\alpha& \sigma \\
	\sigma &\beta
    \end{matrix}\right]u - 2u^\mathrm{T}\left[ \begin{matrix}
	\alpha& \sigma\\
	\sigma &\beta
    \end{matrix}\right]u^r. \\
\label{eq:qp}
\end{equation}

In detail, we define the approaching obstacles as those whose distances to the robot are smaller than a threshold value. With the current position of the robot $(x_0, y_0)$ and the current positions of approaching obstacles $(x_i, y_i), i = 1,2,...,n$, we first predict the next $k$ steps positions of each obstacle $(x^j_i, y^j_i), j = 1,2,...,k$ using constant velocity model. Then we solve the line $l_{\theta}: -\sin(\theta)x+\cos(\theta)y=0$ that maximizes the distance to all approaching obstacles 
\begin{equation}
    \underset{\theta}{\max}\, J(\theta),~~ J(\theta):= \sum_{j=1}^{k}\sum_{i=1}^{n} d^j_i.
\end{equation}
where $d^j_i$ is the distance of shifted obstacle $(x^j_i-x_0, y^j_i-y_0)$ to the line $l_{\theta}$. In this setting, we regard the robot system $(x_0, y_0)$ as the origin and calculate $\theta_{\max}$ that has largest overall distance. With $\theta_{\max}$, we can get eigenvector $\boldsymbol x_1 = (-\sin(\theta_{\max}), \cos(\theta_{\max}))$, which is the safest direction for all approaching obstacles, and $J(\theta_{\max})$ is its corresponding eigenvalue $\lambda_1$. The larger $\lambda_1$ is, the more we want to project the safe control to $\boldsymbol x_1$ direction. The second eigenvector $\boldsymbol x_2$ is perpendicular to $\boldsymbol x_1$ and has smallest overall distance $\lambda_2$. Then, we can build the QP parameter matrix, which is represented as the ellipse in \cref{fig:adapted1} and \cref{fig:adapted2}, as follows:
\begin{equation}
[\boldsymbol x_1, \boldsymbol x_2] \left[\begin{matrix}
	\lambda_1& 0 \\
	0 &\lambda_2
    \end{matrix}\right] [\boldsymbol x_1, \boldsymbol x_2]^{-1} = \left[ \begin{matrix}
	\alpha& \sigma \\
	\sigma &\beta
    \end{matrix}\right]=:Q.
\end{equation}

\subsection{Exploration under Safety Constraints}
In real world reward-sparse and clustered dynamic environments, it's challenging to find a sequence of actions that can lead to positive reward and generalize to related situations, thus RL agents need long training time. Traditional exploration techniques used to address this problem are not suitable for safety-critical tasks, as they may explore unsafe controls. In our framework, with the help of SSA, we add safety constraints to the following two exploration strategies 
to improve the suboptimal policy safely.

\textbf{Parameter Space Noise (PSN)} \cite{DBLP:journals/corr/PlappertHDSCCAA17}\;\;At the start of each episode, we create a copy of RL policy and add noise directly to the policy's parameters, which can lead to consistent exploration and a richer set of behaviors. Suppose we parameterize the policy $\pi_{\theta}$ as a neural network with weights $\theta$. Then the exploration policy is $\pi_{\widetilde{\theta}}$ , where
\begin{equation}
\widetilde{\theta} = \theta + N (0, \sigma^2 I)
\end{equation}

\textbf{Random Network Distillation (RND)} \cite{DBLP:journals/corr/abs-1810-12894}\;\;This exploration strategy will modify the reward function to encourage the agent to visit novel states. In detail, we create two neural networks that take the state $s=(x_R, x^{c}_E)$ as input and train one of the networks to predict the output of the other. The prediction error of two neural networks is defined as novelty and will be added to reward:
\begin{equation}
    \widetilde{r}(s, a)= r(s, a) + ||f_{\theta_1}(s)-f_{\theta_2}(s)||_2^2
\end{equation}

\subsection{Learning from SSA Safe Demonstration}
\begin{figure}
     \centering
     \begin{subfigure}{0.20\textwidth}
         \centering
         \includegraphics[width=1.4\textwidth]{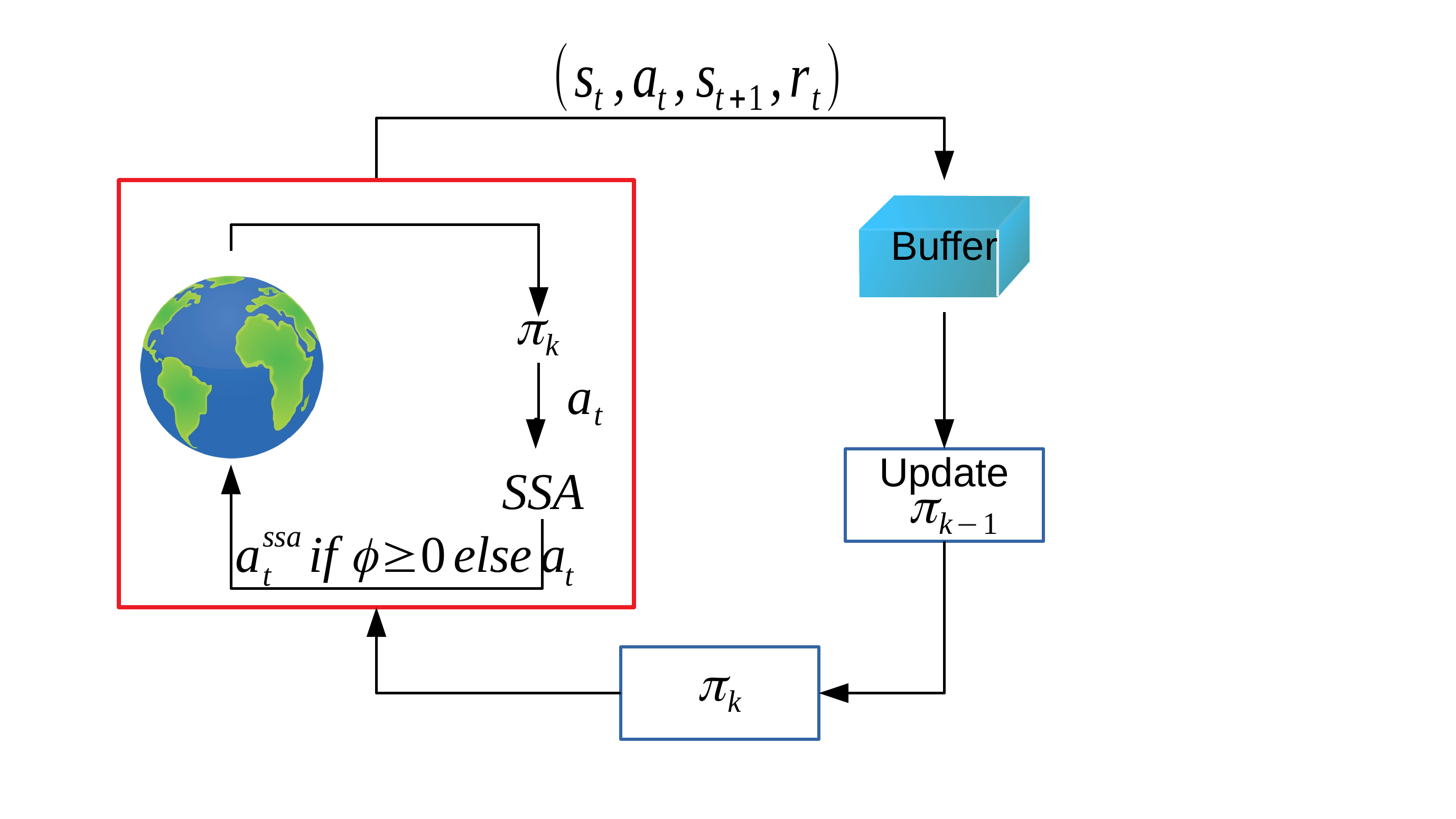}
         \caption{SSA+RL default learning}
         \label{fig:default learning}
     \end{subfigure}
     \begin{subfigure}{0.20\textwidth}
         \centering
         \includegraphics[width=1.4\textwidth]{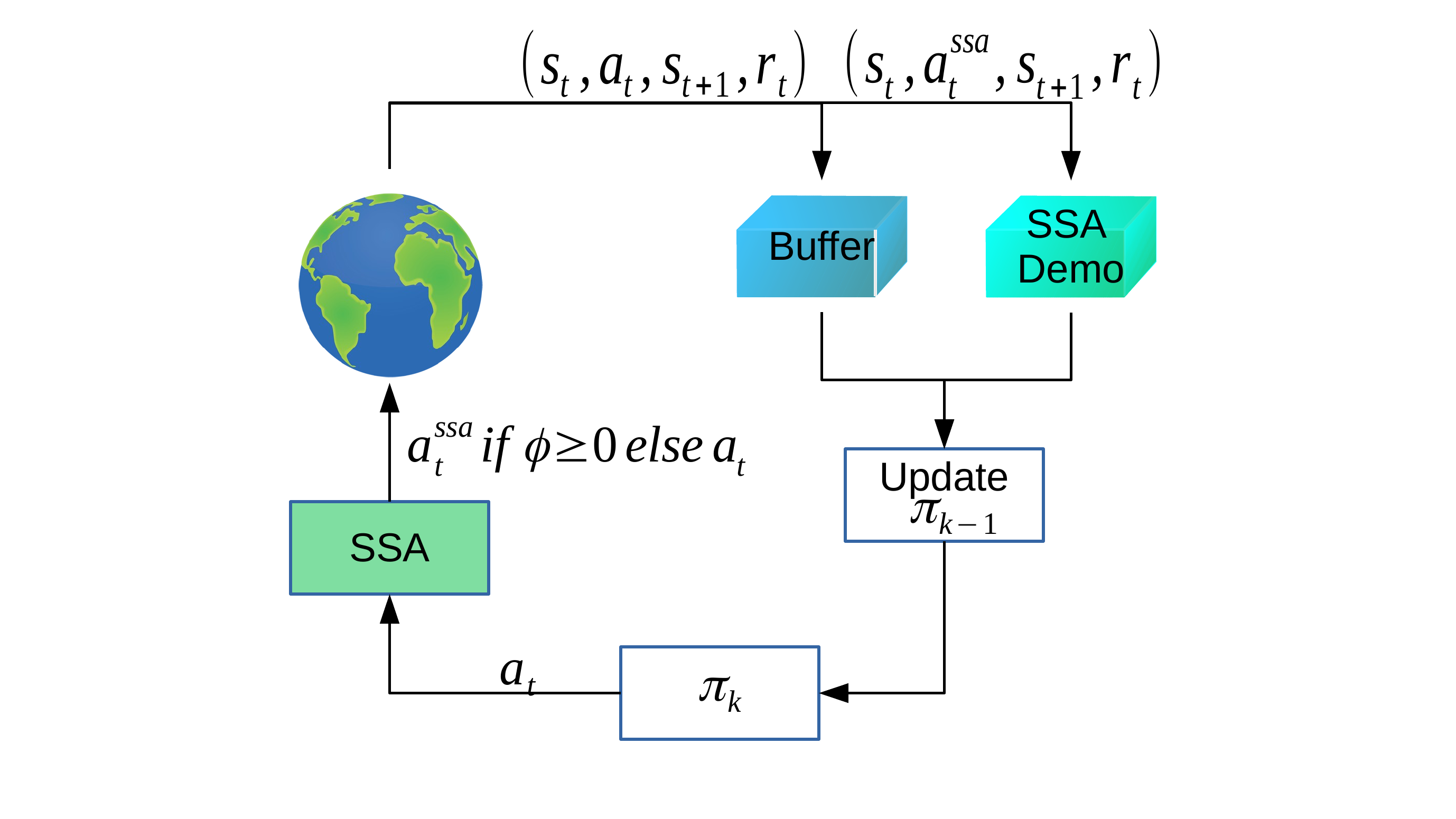}
         \caption{SSA+RL safe learning}
         \label{fig:safe learning}
     \end{subfigure}
     \caption{Block diagrams of the default learning and proposed safe learning. In default learning, the environment (the red box) contains the world and SSA module, while in safe learning, SSA is separated from the environment.}
     \label{fig:three graphs}
     \vspace{-20pt}
\end{figure}

Another technique people use to improve sample efficiency is learning from demonstration (LfD) instead of learning from scratch. Different from the traditional LfD or safe controller guided training in \cite{article}, we don't need to prepare demonstration data and pre-train the RL agent or approximate all prior safe controllers. Instead, SSA would generate safe controls during the interactions with the world and these safe controls are regarded as expert demonstration. To be more specific, in default SSA+RL framework \cref{fig:default learning}, SSA is part of the environment which means the RL agent's control signal will be modified when $\phi \geq 0$ and the agent wouldn't realize that. While in safe learning, we separate the SSA from the environment and make it an independent module, see \cref{fig:safe learning}. In this way, the agent could know the world is taking $a_t$ or $a^{ssa}_t$, then store the self-generated data and demonstration data into two buffers. When updating the policy, we simply use a fixed ratio between self-generated data and SSA demonstrations to mix the training samples. 

\section{Experiments}
\textbf{Environment and Evaluation Method}\;We evaluate our proposed framework in a clustered dynamic environment with sparse reward. Our goal is to move the vehicle, starting from the bottom, to the green area on the top while avoiding 50 moving obstacles in between, which is challenging to be solved by the state-of-the art RL algorithms alone. A success case is shown in \cref{fig:Success_trajectory}. We assume the vehicle can sense the correct positions and velocities of obstacles, but doesn't know their accelerations. The obstacles will be randomly initialized at each episode. This environment tries to simulate the real world scenarios like parking lots and busy streets that have multiple dynamic objects moving around.

We adopt the Twin Delayed Deep Deterministic Policy Gradients (TD3) \cite{DBLP:conf/icml/FujimotoHM18} as our baseline RL model. For experiments that evaluate safety, we train the models for 50 episodes as we notice it's long enough for the SSA-based models to converge. For experiments that evaluate sample efficiency, we train the models until their performances reach a threshold reward $R_{min}$. The required $R_{min}$ is to achieve at least 1900 on average for the past 20 episodes. For models that fail to meet $R_{min}$ within 1000 episodes, we set its result as 1000 episodes. We repeat each training with different seeds for 10 times and calculate the average performance. The code is open sourced at https://github.com/hychen-naza/SSA-RL.

\textbf{Hypothesis}\;we evaluate the proposed framework by verifying the following four hypotheses:
\begin{itemize}
    \item H1: The SSA+RL framework can greatly reduce safety violation comparing to the TD3 baseline model in clustered dynamic environment.
    \item H2: The adapted SSA can achieve better efficiency and higher task success rate than vanilla SSA.
    \item H3: Traditional exploration strategies are not safe and hence not data efficient, but exploration under safety constraints could improve the sample efficiency.
    \item H4: Direct learning from the safe controls  demonstrated by SSA can best speedup training and maintain safety compared to pure reward-driven approaches.
    \item H5: Compared to other safe RL methods, our SSA+RL framework can achieve best safety performance.
\end{itemize}

\subsection{Results}
\begin{figure*}[htbp]
     \centering
     \begin{subfigure}{0.4\textwidth}
         \centering
         \includegraphics[width=\textwidth, height=4cm]{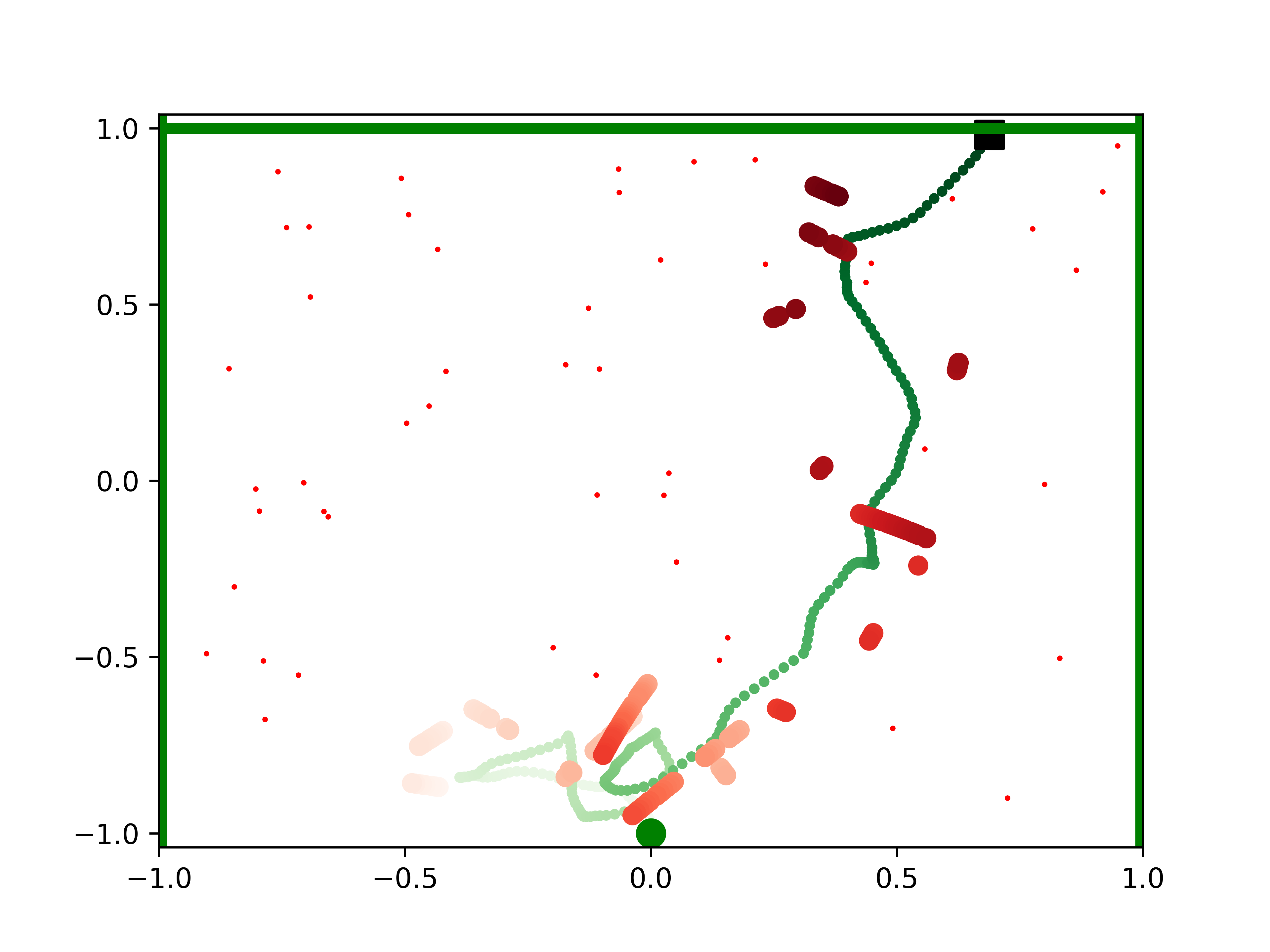}
         \caption{Success case}
         \label{fig:Success_trajectory}
     \end{subfigure}
     \begin{subfigure}{0.4\textwidth}
         \centering
         \includegraphics[width=\textwidth, height=4cm]{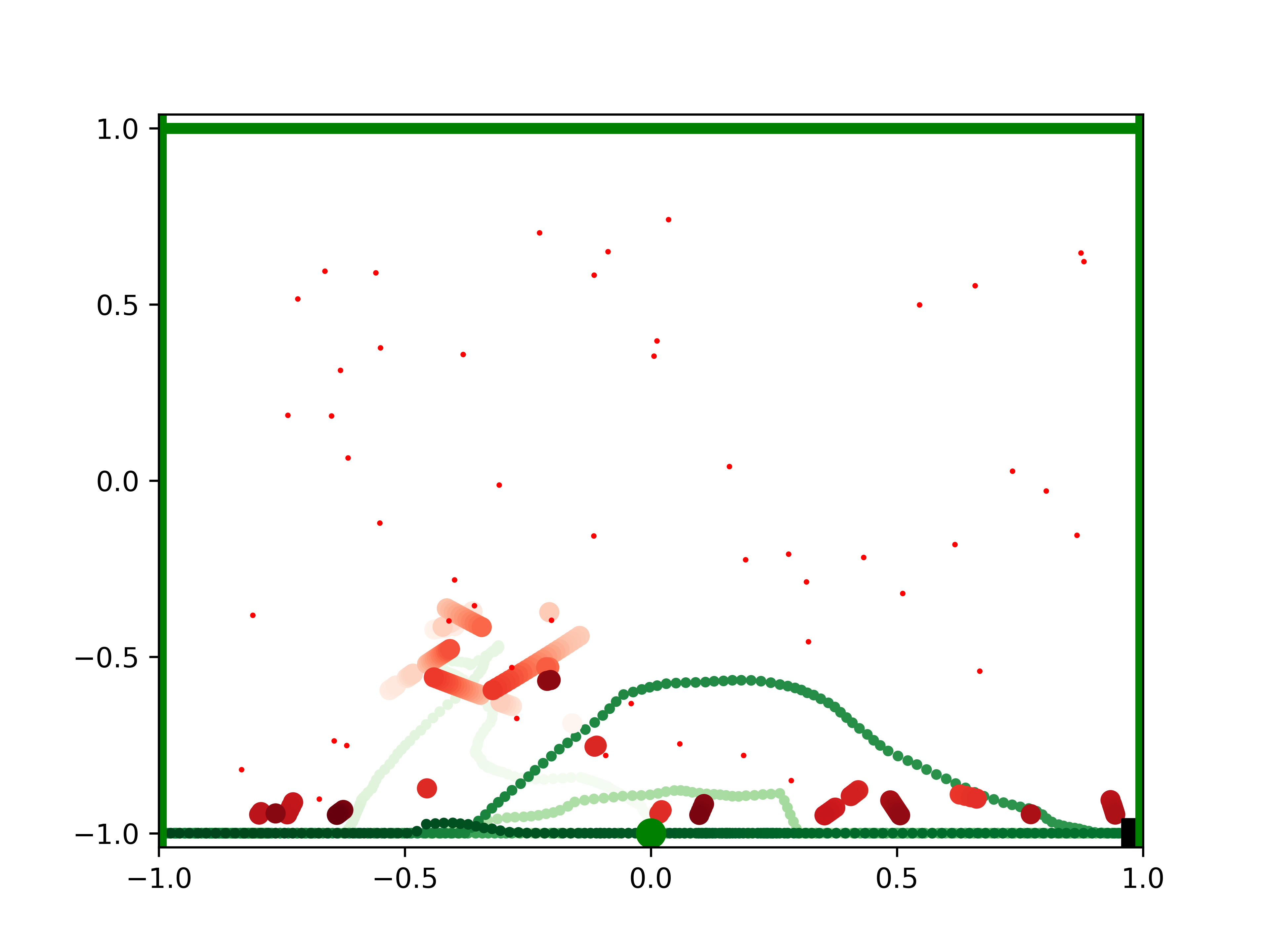}
         \caption{Failure case}
         \label{fig:Failure_trajectory}
     \end{subfigure}
     
     \begin{subfigure}{0.4\textwidth}
         \centering
         \includegraphics[width=\textwidth, height=4cm]{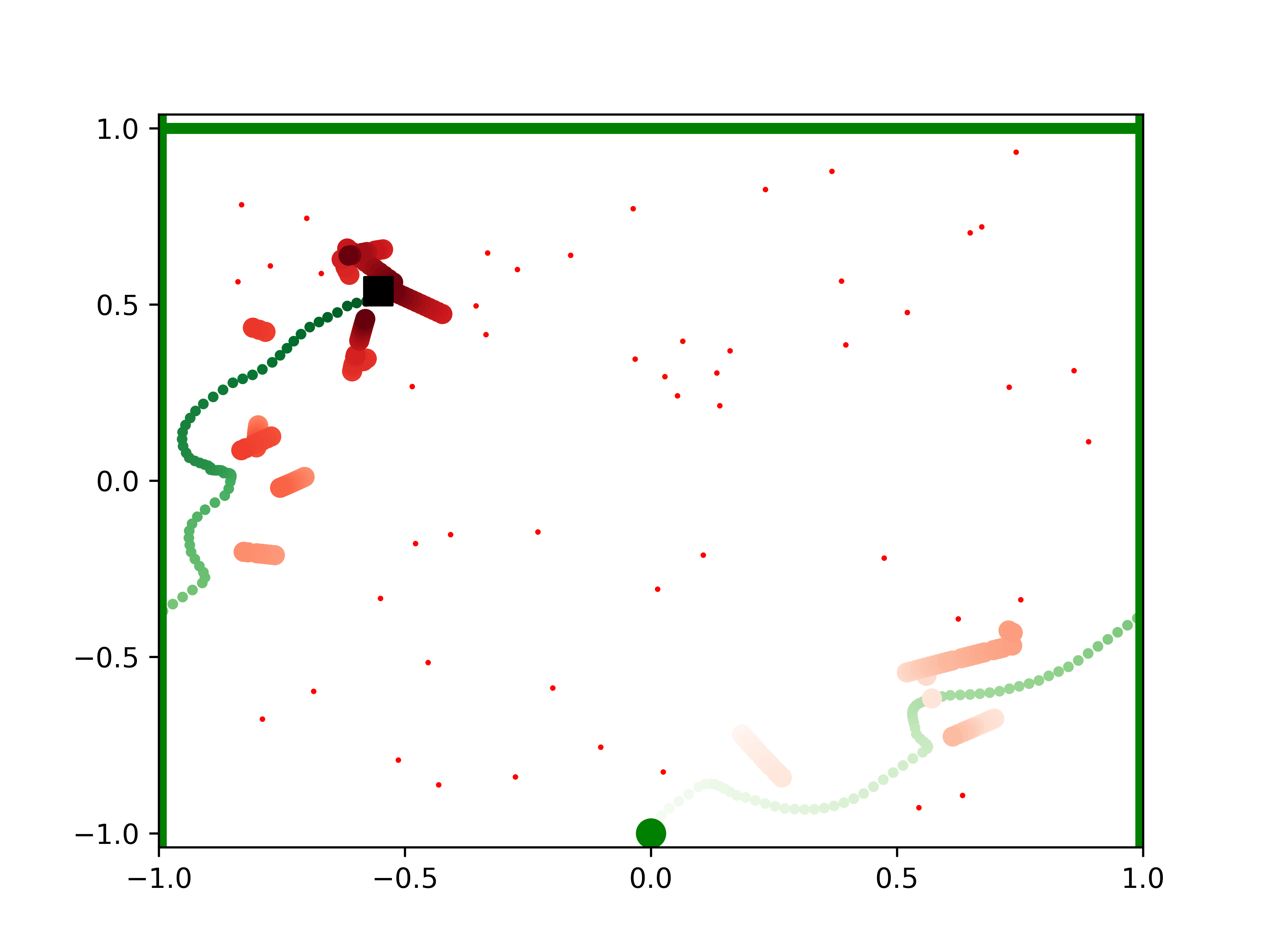}
         \caption{Inevitable collision case}
         \label{fig:Collision_trajectory}
     \end{subfigure}
     \begin{subfigure}{0.4\textwidth}
         \centering
         \includegraphics[width=\textwidth, height=4cm]{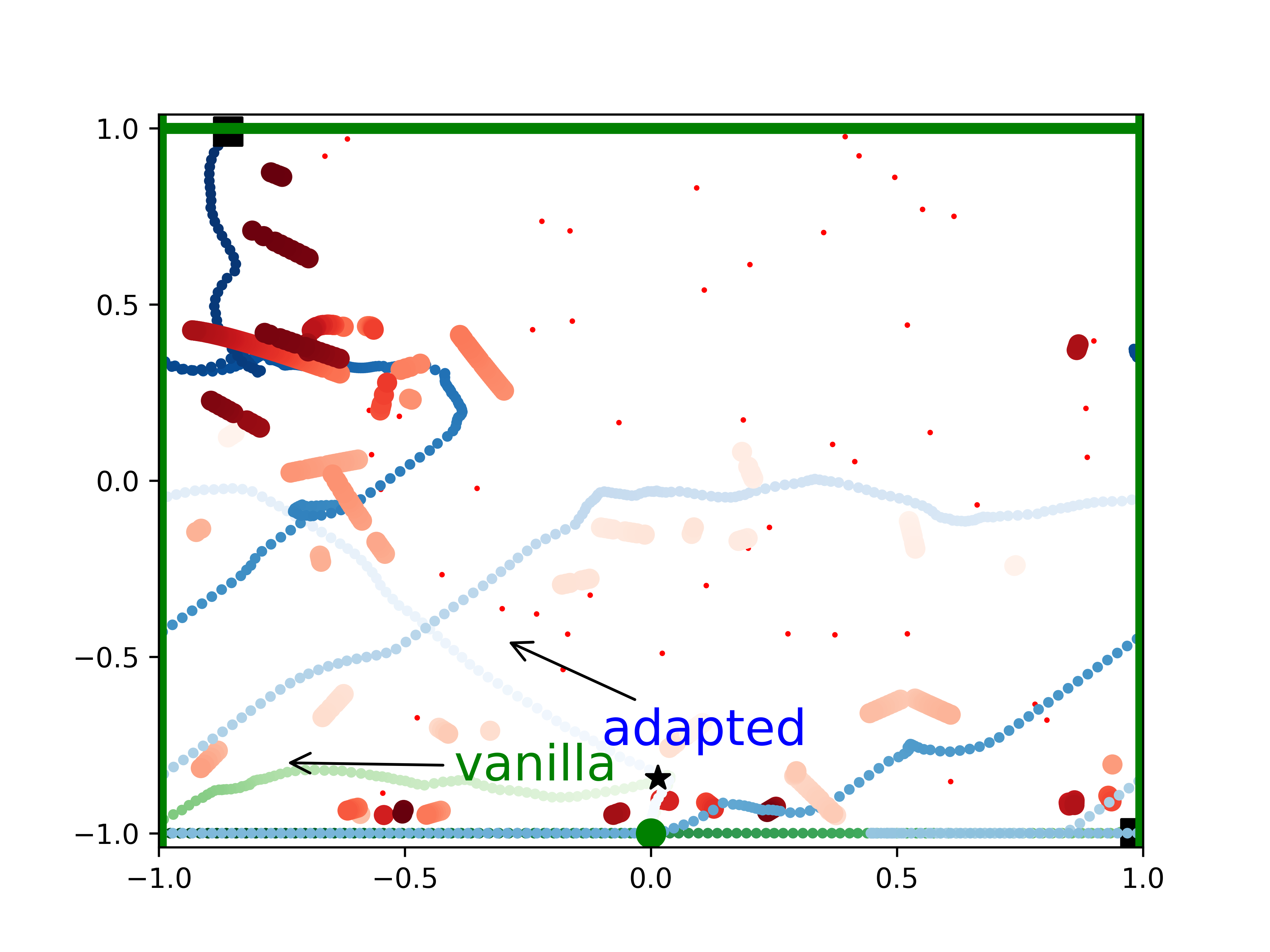}
         \caption{Adapted SSA (blue) and Vanilla SSA (green).}
         \label{fig:Adapted_SSA_trajectory}
     \end{subfigure}

     \begin{subfigure}{0.8\textwidth}
         \centering
         \includegraphics[width=\textwidth]{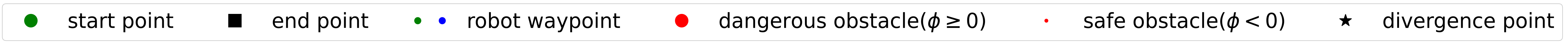}
         \label{fig:legend}
     \end{subfigure}
     \vspace{-15pt}
     \caption{Trajectory plots of four cases. For both vehicle and obstacles, the darker color means more recent position while the lighter color means older position. The vertical green lines are boundaries that will wrap around. The top green line represents the goal area.}
     \label{fig:three graphs}
     \vspace{-15pt}
\end{figure*}

\begin{figure*}[htbp]
     \centering
     \begin{subfigure}{0.23\textwidth}
         \centering
         \includegraphics[width=\textwidth]{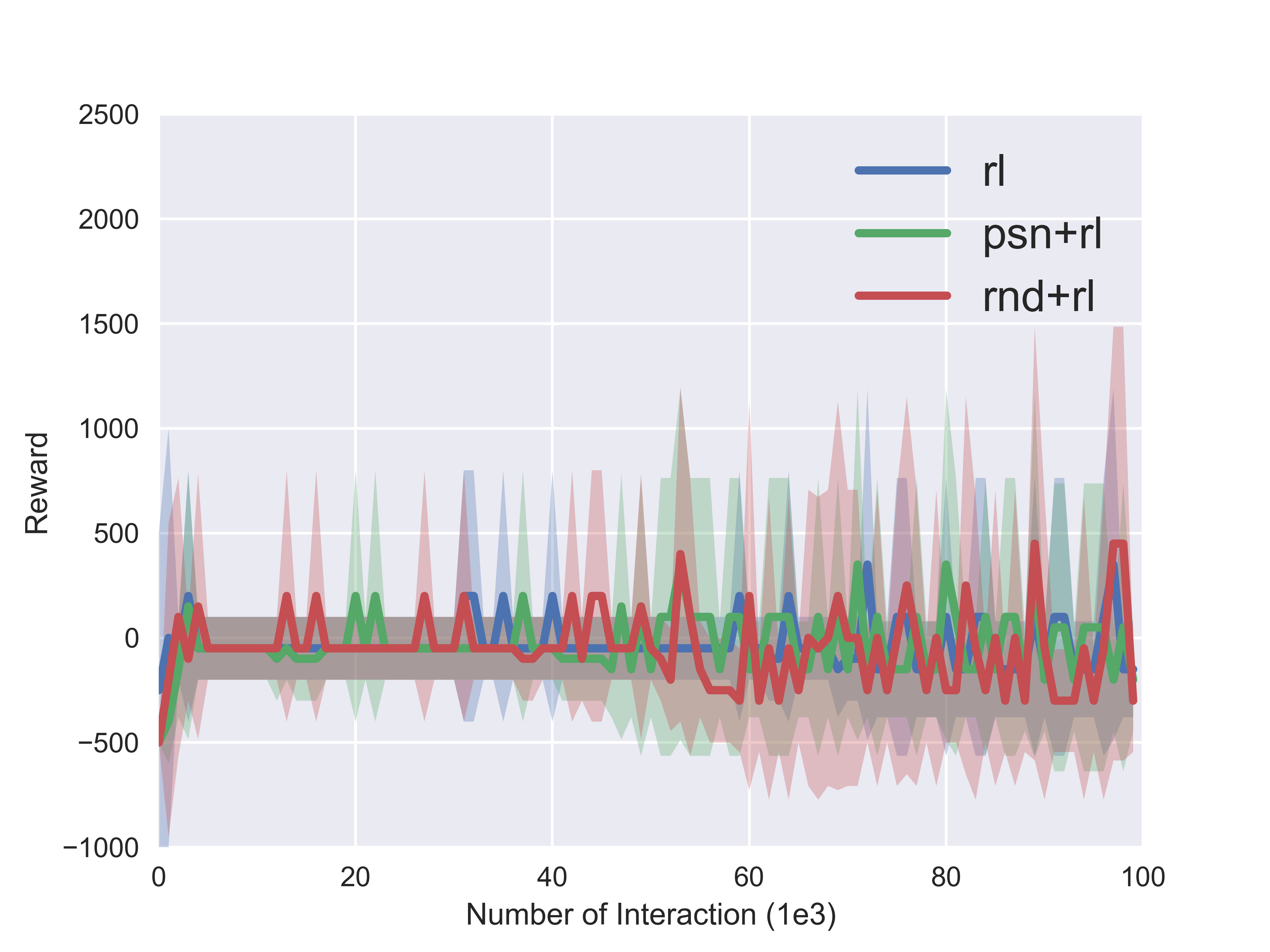}
         \caption{Baseline models}
         \label{fig:plot_qp}
     \end{subfigure}
     \begin{subfigure}{0.23\textwidth}
         \centering
         \includegraphics[width=\textwidth]{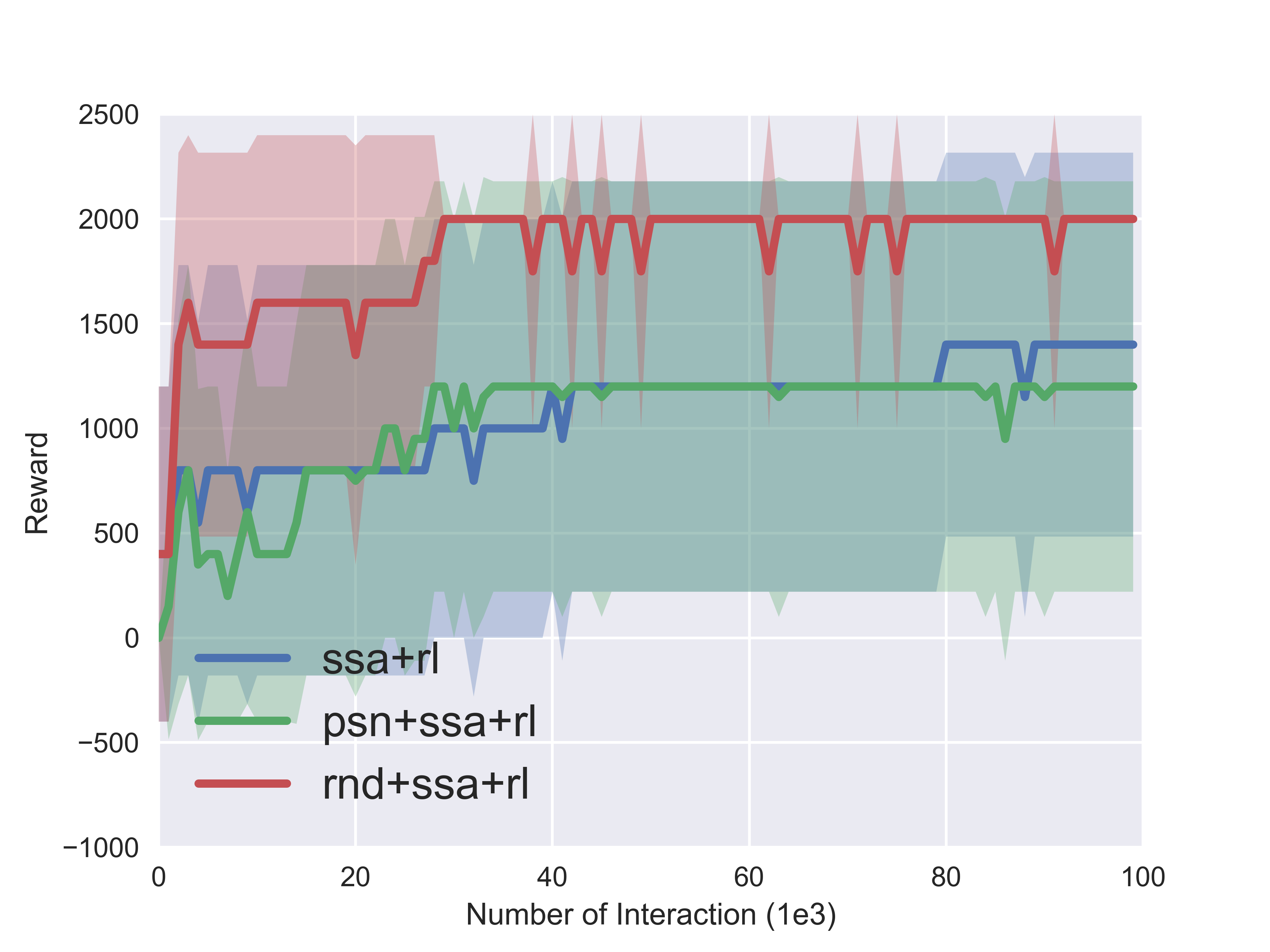}
         \caption{Exploration with SSA}
         \label{fig:plot_exploration_ssa}
     \end{subfigure}
     \begin{subfigure}{0.23\textwidth}
         \centering
         \includegraphics[width=\textwidth]{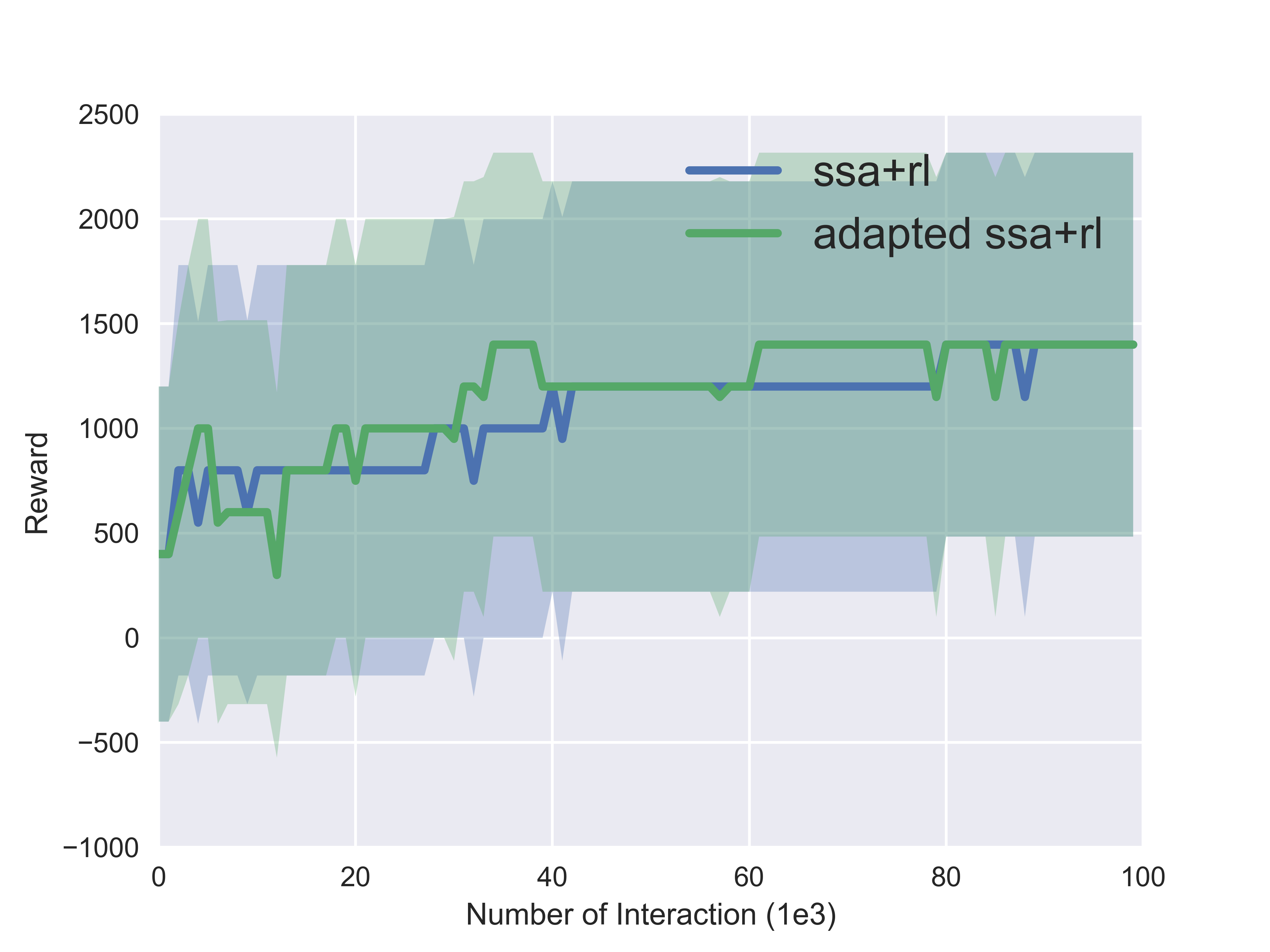}
         \caption{Adapted SSA}
         \label{fig:plot_exploration}
     \end{subfigure}
     \begin{subfigure}{0.23\textwidth}
         \centering
         \includegraphics[width=\textwidth]{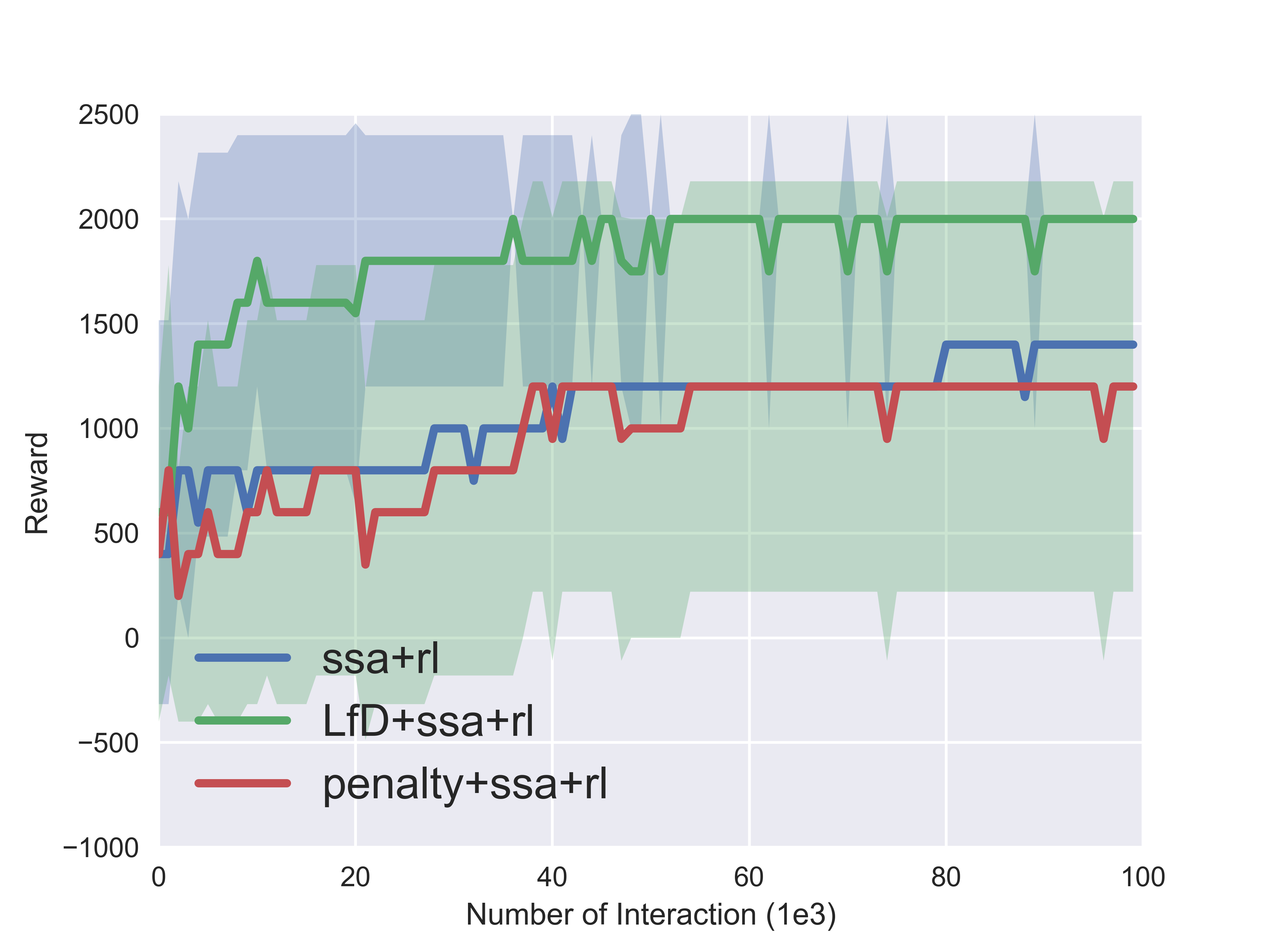}
         \caption{Learning from SSA}
         \label{fig:plot_learning}
     \end{subfigure}
     \caption{Average performance of baseline models and our proposed models over $10^6$ steps interactions.}
     \label{fig:three graphs}
     \vspace{-15pt}
\end{figure*}

\begin{figure}[htbp]
    \vspace{-5pt}
     \centering
     \begin{subfigure}{0.21\textwidth}
         \centering
         \includegraphics[width=1.1\textwidth]{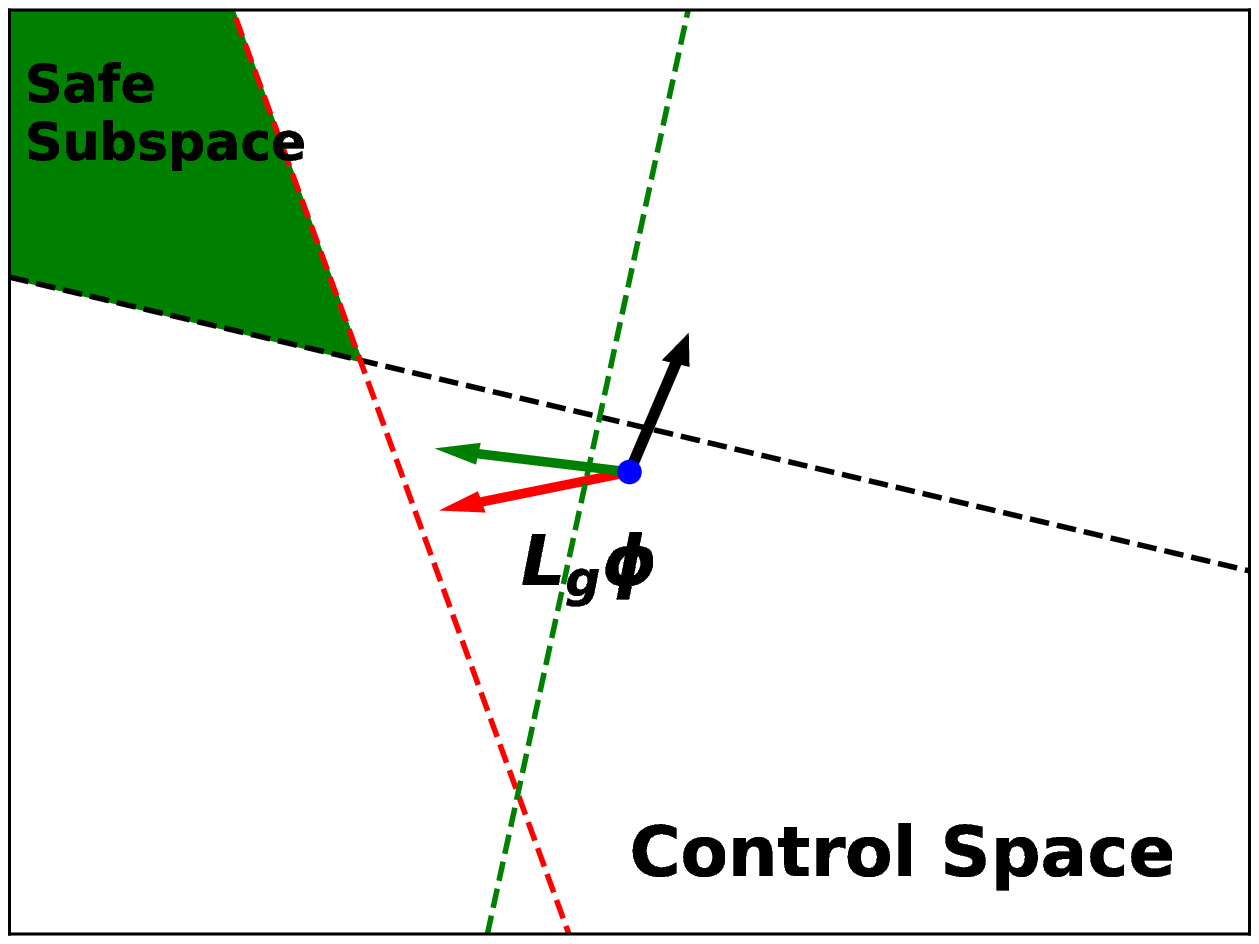}
         \caption{Safe subspace exists}
         \label{fig:safe control subspace}
     \end{subfigure}
     \begin{subfigure}{0.21\textwidth}
         \centering
         \includegraphics[width=1.1\textwidth]{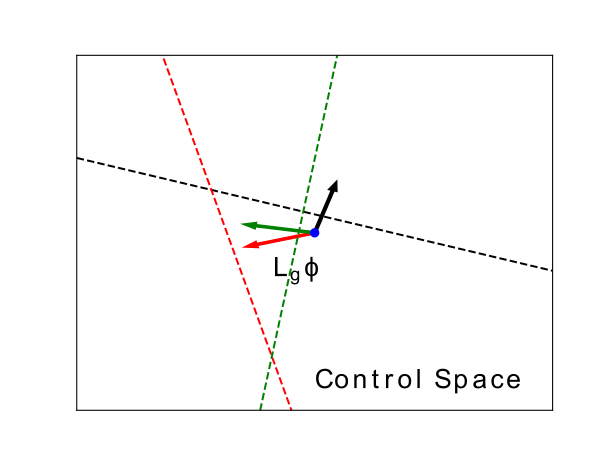}
         \caption{No safe control exists}
         \label{fig:no safe control learning}
     \end{subfigure}
     \caption{SSA feasibility when meeting multiple obstacles.}
     \label{fig:feasibility}
     \vspace{-25pt}
\end{figure}

\textbf{H1:}\;Through our experiments, TD3 model gets 31.7\% collision rate and 66.3\% failure rate on average, see \cref{tab:safety result}. During training, the policy gradually converges to a local optimum after repeated collisions, which is staying at the bottom where obstacles can't get to. On the other hand, SSA helps to reduce the collision rate from $31.7\%$ to $0.8\%$. We find SSA can't achieve 0-safety violation discussed in other papers using static testing cases like avoiding fixed obstacles or static hazard areas; in which cases, these always exist a safe control to meet safety constraints \cite{article}\cite{zhao2021modelfree}. But in our environment, the obstacles are dynamic and the vehicle can be surrounded by multiple obstacles. In \cref{fig:Collision_trajectory}, there were four obstacles driving from three different directions toward the vehicle, and SSA cannot find a safe control meet all safety constraints and collision happens (as shown in the control space plot in \cref{fig:no safe control learning}).
Besides, the success rate increased from $2\%$ to $50.2\%$ as SSA prevents the RL agent from collisions and increases the possibility of reaching the goal. But SSA+RL may still stuck in local optimums as its failure rate is 49\%. In  \cref{fig:Failure_trajectory}, the vehicle tries to navigate towards the goal, but it is pushed back by obstacles and stays at the bottom to avoid collision. 

\begin{table}[htbp]
\vspace{-8pt}
\fontsize{7.5}{9}\selectfont
\centering
\caption {Safety comparison between proposed models.}
\label{tab:safety result}
\begin{tabular}{@{}cccccc@{}}
\toprule
 & Model & Success & Failure & Collision & Reward\\ 
 \midrule
 \multirow{3}*{\makecell[c]{Baseline\\Models}} 
&{RL} & 2\%& 66.3\% & 31.7\% & -118.6             \\
&{PSN+RL} & 3.6\% & 76.8\% & 19.6\%&-26\\
&{RND+RL} & 5.2\% & 49.5\% & 47.8\%&-133.8\\\midrule
\multirow{5}*{\makecell[c]{Proposed\\Models}} 
&{SSA+RL} & 50.2\% & 49\% & \textbf{0.8\%}&1000\\
&{Adapted SSA+RL} & 68.6\%& 30\% & 1.4\% & 1365\\
&{PSN+SSA+RL} & 40.4\% & 58.3\% & 1.3\%&800\\
&{RND+SSA+RL} & 71.4\% & 27.2\% & 1.4\%&1421\\
&{LfD+SSA+RL}& \textbf{89.8\%} & \textbf{9.4\%} & \textbf{0.8\%} & \textbf{1792}\\
&{Penalty+SSA+RL} & 43.8\% & 55.2\% & 1.0\%&871\\
\bottomrule
\end{tabular}
\vspace{-8pt}
\end{table}

\textbf{H2:}\;The improvement of SSA adaptation is less significant compared to other techniques we used due to two main reasons. Firstly, SSA adaption only works when SSA is triggered, which wouldn't happen at every step. Secondly, in many cases, vanilla SSA and adapted SSA will generate very close safe controls as both of them need to satisfy the same hard safety constraint. As shown in \cref{tab:efficiency result}, the averaged interaction number of adapted SSA+RL model is $2.3\times10^5$, which is very close to the interaction number $2.4\times10^5$ of SSA+RL model, because the adapted SSA+RL model may also get stuck in the local optimum. Nevertheless, we find the failure rate drops from $49\%$ to $30\%$ and the success rate increases from $50.2\%$ to $68.6\%$ in \cref{tab:safety result}. Because adapted SSA generates fewer detours when meeting dangerous obstacles, and the projection guidance leads the vehicle to a more efficient direction. The effects of adapted safe controls will accumulate and improve the overall success rate and reward. In \cref{fig:Adapted_SSA_trajectory}, adapted SSA and vanilla SSA are tested in the same environment. At the beginning, their paths are identical, but diverge gradually. When meeting dangerous obstacle at the black star point, which is the most critical divergence point, the adapted SSA outputs control in upwards direction while the vanilla SSA outputs control in downwards direction. At last, the adapted SSA helps the vehicle reach the goal while the vanilla SSA pushes the vehicle back to the bottom.

\begin{table}[htbp]
\vspace{-7pt}
\fontsize{8}{9}\selectfont
\centering
\caption {Efficiency comparison between proposed models.}
\label{tab:efficiency result}
\begin{tabular}{@{}cccc@{}}
\toprule
 & Model & Episodes & Interaction \\ 
 \midrule
 \multirow{3}*{\makecell[c]{Baseline\\Models
}} 
&{RL} & 1000& $10^6$           \\
&{PSN+RL} & 1000& $10^6$\\
&{RND+RL} & 1000& $10^6$\\ \midrule
\multirow{5}*{\makecell[c]{Proposed\\Models}} 
&{SSA+RL} & 254 & 2.4$\times10^5$\\
&{Adapted SSA+RL} & 245 & 2.3$\times10^5$\\
&{PSN+SSA+RL} & 251 & 2.4$\times10^5$\\
&{RND+SSA+RL} & \textbf{23} & \textbf{6183}\\
&{LfD+SSA+RL}& 26 & 8464\\
&{Penalty+SSA+RL}& 331 &3.2$\times10^5$\\
\bottomrule
\end{tabular}
\vspace{-7pt}
\end{table}

\textbf{H3:}\;This hypothesis can be proved via results in \cref{tab:safety result}. The exploration strategies PSN and RND only improve the success rate slightly from 2\% to 3.6\% and 5.2\% respectively. But the failure rate of PSN-enabled model increases to 76.8\% and the collision rate of RND-enabled model increases 47.8\%. That is mainly because these exploration strategies have large portion of harmful exploration, i.e., unsafe controls that don't meet the safety constraint $\dot{\phi} \leq -\eta\,\phi$. The RND+SSA+RL model can reduce the failure rate to 27.2\% and increase the success rate to 71.4\%, which mitigates the problem of being overly conservative in clustered environment. Moreover, in \cref{tab:efficiency result}, this model uses substantially fewer episodes and 38$\times$ fewer interactions to meet $R_{min}$ compared to the SSA+RL model. By visiting new states, the agent has more chance to avoid getting stuck in the local optimum and converge to optimal policy faster. However, PSN+SSA+RL gets worse performance than SSA+RL, which may because PSN is unable to sufficiently explore in our challenging environments. 

\vspace{-1pt}

\textbf{H4:}\;For LfD+SSA+RL, it takes only 26 episodes and 8464 steps to meet $R_{min}$, using 28$\times$ fewer steps compared to SSA+RL model in \cref{tab:efficiency result}. From training plot \cref{fig:plot_learning}, this model learns the optimal controller in all experiments, achieving faster convergence and greater training stability than other models. Moreover, LfD+SSA+RL achieves highest success rate 89.8\% and lowest collision rate 0.8\% in \cref{tab:safety result}, which proves that learning from SSA demonstration can best maintain safety. This is due to two reasons: Firstly, there is a mismatch between the control that RL agent generates and the real control that environment takes due to control modification in the default SSA+RL framework. This introduces errors to the training data and lower the training efficiency. Secondly, the RL policy could better learn how to take safe control with SSA demonstration, instead of learning from scratch or the reward penalty. To validate the second point, we give the agent a negative reward penalty $- ||a^{ssa}_t - a_t||_2^2$ to negatively reinforce unsafe control following the idea in \cite{10.5555/3237383.3238074}. The results show that penalty+SSA+RL model has lower success rate 43.8\%, slightly higher collision rate 1.0\% and takes more interactions steps compared to SSA+RL model in \cref{tab:safety result} and \cref{tab:efficiency result}. This is because learning safe control from the penalty is too hard for the agent, which reduces the learning efficiency. 

\textbf{H5}\; We compare our results with the recent safe RL methods from three categories: constrained policy optimization (CPO) \cite{DBLP:journals/corr/AchiamHTA17}, probabilistic shields+RL \cite{Jansen2018ShieldedDI}, CBF+RL \cite{article}. CPO will converges to constraint-satisfying policies in the end, but  not consistently constraint-satisfying throughout training. Probability shield is applied in discrete MDPs, in which they can calculate the safety violation probability for all possible actions and states within the finite horizon. But this method is hard to check all possible actions in continuous systems and will inevitably be suboptimal. The CBF used in \cite{article} is only tested in simple environments with linear barrier function $h(x) = p^{T}x+q$. Moreover, the CBF approach in general has limitations than the SSA approach: 1) enforce control constraint everywhere, which is not necessary; 2) our SSA method uses a design rule \cite{liu2014control}\cite{zhao2021modelfree} to nonlinearly synthesize $\phi$ from $\phi_0$ which ensures that there always exists a feasible control input to satisfy the safety constraint on control, while the counterpart in CBF is the exponential CBF which is still a linear function on $\phi_0$. Nonetheless, the design rule can also be applied to CBF and we added a comparison of CBF+RL using the same $\phi$. In \cref{tab:safety comparison result}, CPO gets high collision rate since it still relies on trial-and-error to enforce constraints. Probabilistic Shields+RL can only guarantee safety with some probability and is too restrictive to the agent. The barrier function in CBF+RL is too simple to correctly evaluate the safety in higher dimensional spaces. The collision rate of CBF using safety index $\phi$ is 1.0\% higher than that of SSA+RL. The reason is that CBF adds constraints for all nearby obstacles even for safe ones, which restrict the control space and may fail to find safe control for dangerous obstacles.

\begin{table}[htbp]
\vspace{-7pt}
\fontsize{7.5}{9}\selectfont
\centering
\caption {Safety comparison with other methods.}
\label{tab:safety comparison result}
\begin{tabular}{@{}ccccc@{}}
\toprule
 Model & Success & Failure & Collision & Reward\\ 
 \midrule
{CPO} & 7.4\% & 6.8\% & 85.8\%&-281\\
{Probabilistic Shields+RL} & 13.8\% & 80.4\% & 5.8\%&287\\
{CBF+RL} & 74.6\% & 21.8\% & 3.6\%&1474\\
{CBF+Safety Index $\phi$+RL} & 52.8\% & 45.4\% & 1.8\%&1047\\
\bottomrule
\end{tabular}
\vspace{-15pt}
\end{table}

\section{Conclusion}
In this work, we propose to use SSA to improve safety during RL policy training, and introduce three strategies, including SSA adaption, exploration under safety constraint and learning from SSA demonstrations, to improve the learning efficiency. We further validate the proposed framework in a clustered dynamic environment. The results show that SSA can greatly reduce the safety-violation except for the situations that no safe control exists, and achieve better safety performance statistically compare to other safe RL methods. Combining with the three strategies, the agent can solve the task with substantially fewer episodes and interactions.

\bibliographystyle{IEEEtran}
\bibliography{lcsys.bib}

\end{document}